\documentclass{llncs}

\usepackage{amstext,amsmath,latexsym}
\usepackage{epsfig}
\usepackage{verbatim}
\usepackage{pslatex}

\newcommand{\NID}{ \textsc {NID} }
\newcommand{\NGD}{ \textsc {NGD} }

\newcommand{\NCD}{ \textsc {NCD} }

\newcommand{\SVM}{ \textsc {SVM} }

\begin{document}
\pagestyle{empty}

\mainmatter

\title{Similarity of Objects and the Meaning of Words\thanks{This work supported in part
by the EU sixth framework project RESQ, IST--1999--11234,
the NoE QUIPROCONE IST--1999--29064,
the ESF QiT Programmme, and 
the EU NoE PASCAL, and by the Netherlands Organization for
Scientific Research (NWO) under Grant 612.052.004.
}
}
\titlerunning{Lecture Notes in Computer Science}
\author{Rudi Cilibrasi 
and
Paul Vit\'{a}nyi
}

\authorrunning{Rudi Cilibrasi and Paul Vit\'{a}nyi}

\institute{
CWI, Kruislaan 413, 1098 SJ Amsterdam,\\
The Netherlands. Email: Rudi.Cilibrasi@cwi.nl;  Paul.Vitanyi@cwi.nl
}


\author{Rudi Cilibrasi
\and 
Paul Vitanyi\thanks{Also affiliated with the Computer Science 
Department, University of Amsterdam.
}}

\maketitle

\begin{abstract}
We survey the emerging area of compression-based, parameter-free, 
similarity distance measures
useful in data-mining,
pattern recognition, learning and automatic semantics extraction.
Given a family of distances on a set of objects,
a distance is universal up to a certain precision for that family if it
minorizes every distance in the family between every two objects 
in the set, up to the stated precision (we do not require the universal
distance to be an element of the family).
We consider similarity distances 
for two types of objects: literal objects that as such contain all of their
meaning, like genomes or books, and names for objects.
The latter may have
literal embodyments like the first type, but may also
be abstract like ``red'' or ``christianity.'' For the first type
we consider 
a family of computable distance measures
corresponding to parameters expressing similarity according to
particular features
between 
pairs of literal objects. For the second type we consider similarity
distances generated by web users corresponding to particular semantic
relations between the (names for) the designated objects. 
For both families we give universal similarity
distance measures, incorporating all particular distance measures
in the family. In the first case the universal
distance is based on compression and in the second
case it is based on Google page counts related to search terms.
In both cases experiments on a massive scale give evidence of the
viability of the approaches.
\end{abstract}

\section{Introduction}
Objects can be given literally, like the literal
four-letter genome of a mouse,
or the literal text of {\em War and Peace} by Tolstoy. For
simplicity we take it that all meaning of the object
is represented by the literal object itself. Objects can also be
given by name, like ``the four-letter genome of a mouse,''
or ``the text of {\em War and Peace} by Tolstoy.'' There are
also objects that cannot be given literally, but only by name
and acquire their meaning from their contexts in background common
knowledge in humankind, like ``home'' or ``red.''
In the literal setting, objective similarity of objects can be established
by feature analysis, one type of similarity per feature.
In the abstract ``name'' setting, all similarity must depend on 
background knowledge and common semantics relations,
which is inherently subjective and ``in the mind of the beholder.'' 

\subsection{Compression Based Similarity}
All data are created equal but some data are more alike than others.
We and others have recently proposed very general
methods expressing this alikeness,
using a new similarity metric based on compression.
It is parameter-free in that it
doesn't use any features or background knowledge 
about the data, and can without
changes be applied to different areas and across area boundaries.
Put differently: just like `parameter-free' statistical methods,
the new method uses essentially unboundedly many parameters,
the ones that are appropriate.
It is universal in that it approximates the parameter
expressing similarity of the dominant feature in all pairwise
comparisons.
It is robust in the sense that its success appears independent
from the type of compressor used.
The clustering we use is hierarchical clustering in dendrograms
based on a new fast heuristic for the quartet method.
The method is available as an open-source software tool, \cite{Ci03}.

{\bf Feature-Based Similarities:}
We are presented with unknown data and
the question is to determine the similarities among them
and group like with like together. Commonly, the data are
of a certain type: music files, transaction records of ATM machines,
credit card applications, genomic data. In these data there are
hidden relations that we would like to get out in the open.
For example, from genomic data one can extract
letter- or block frequencies (the blocks are over the four-letter alphabet);
 from music files one can extract
various specific numerical features,
related to pitch, rhythm, harmony etc.
One can extract such features using for instance
Fourier transforms~\cite{TC02} or wavelet transforms~\cite{GKCwavelet},
to quantify parameters expressing similarity.
The resulting vectors corresponding to the various files are then
classified or clustered using existing classification software, based on
various standard statistical pattern recognition classifiers~\cite{TC02},
Bayesian classifiers~\cite{DTWml},
hidden Markov models~\cite{CVfolk},
ensembles of nearest-neighbor classifiers~\cite{GKCwavelet}
or neural networks~\cite{DTWml,Sneural}.
For example, in music one feature would be to look for rhythm in the sense
of beats per minute. One can make a histogram where each histogram
bin corresponds to a particular tempo in beats-per-minute and
the associated peak shows how frequent and strong that
particular periodicity was over the entire piece. In \cite{TC02}
we see a gradual change from a few high peaks to many low and spread-out
ones going from hip-hip, rock, jazz, to classical. One can use this
similarity type to try to cluster pieces in these categories.
However, such a method requires specific and detailed knowledge of
the problem area, since one needs to know what features to look for.

{\bf Non-Feature Similarities:}
Our aim
is to capture, in a single similarity metric,
{\em every effective distance\/}:
effective versions of Hamming distance, Euclidean distance,
edit distances, alignment distance, Lempel-Ziv distance,
and so on.
This metric should be so general that it works in every
domain: music, text, literature, programs, genomes, executables,
natural language determination,
equally and simultaneously.
It would be able to simultaneously detect {\em all\/}
similarities between pieces that other effective distances can detect
seperately.

The normalized
version of the ``information metric'' of \cite{liminvit:kolmbook,BGLVZ}
fills the requirements for such a ``universal'' metric. 
Roughly speaking, two objects are deemed close if
we can significantly ``compress'' one given the information
in the other, the idea being that if two pieces are more similar,
then we can more succinctly describe one given the other.
The mathematics used is based on Kolmogorov complexity 
theory \cite{liminvit:kolmbook}. 

\subsection{A Brief History}
In view of the success of the method, in numerous applications,
it is perhaps useful
to trace its descent in some detail. Let $K(x)$ denote
the unconditional Kolmogorov complexity of $x$, and let
$K(x|y)$ denote the conditional Kolmogorov complexity
of $x$ given $y$. Intuitively, the Kolmorov complexity of an object
is the number of bits in the ultimate 
compressed version of the object,
or, more precisely, from which the object can be recovered by a
fixed algorithm.  The ``sum'' version of information
distance, $K(x|y)+K(y|x)$, arose from thermodynamical considerations
about reversible computations \cite{LiVi92,LiVi96} in 1992. 
It is a metric and minorizes all computable
distances satisfying a given density condition
 up to a multiplicative factor of 2. Subsequently, 
in 1993, the ``max''
version of information distance, 
$\max \{K(x|y),K(y|x)\}$, was introduced
in \cite{BGLVZ}. Up to a logarithmic additive
term, it is the length of the shortest binary program
that transforms $x$ into $y$, and $y$ into $x$. 
It is a metric as well, and this metric
minorizes all computable distances satisfying a given
density condition up to an
additive ignorable term. 
This is optimal. But the Kolmogorov complexity
is uncomputable, which seems to preclude application altogether.
However, in 1999 the normalized version of the ``sum''
information distance $(K(x|y)+K(y|x))/K(xy)$ 
was introduced as a similarity
distance and applied to construct a phylogeny of bacteria 
in \cite{CKL99},
and subsequently mammal phylogeny in 2001 \cite{LBCKKZ01},
followed by plagiarism detection
in student programming assignments \cite{SID}, and phylogeny of
chain letters in \cite{BLM03}. In  \cite{LBCKKZ01}
it was shown that the normalized 
sum distance is a metric, and minorizes 
certain computable distances up to a multiplicative factor of 2
with high probability.
In a bold move, in these papers the uncomputable Kolmogorov complexity
was replaced by an approximation using a real-world compressor, for
example the special-purpose genome compressor GenCompress. 
Note that, because of the
uncomputability of the Kolmogorov complexity, in principle
one cannot determine the degree of accuracy of the approximation 
to the target value. Yet it turned out that this practical approximation,
imprecise though it is,
but guided by an ideal
provable theory, in general gives good results on natural data sets. 
The early use of the ``sum'' distance was
replaced by the ``max'' distance in \cite{LiVi01} in 2001 and
applied to mammal phylogeny in 2001 in the early version 
of \cite{LCLMV03} 
and in later versions also to the language tree.
In \cite{LCLMV03}
it was shown that an appropriately
normalized ``max'' distance is metric, and minorizes all
normalized computable distances satisfying a certain density property
up to an additive vanishing term.
That is, it
discovers all effective similarities 
of this family in the sense that if two
objects are close according to some effective similarity, then 
they are also close according to the normalized information distance.
Put differently, the normalized information distance represents
similarity according to the dominating shared feature between
the two objects being compared.
In comparisons of more than two objects, 
different pairs may have different dominating features.
For every two objects,
this universal metric distance zooms in on the dominant
similarity between those two objects
 out of a wide class of admissible similarity
features. 
Hence it may be called {\em ``the'' similarity metric}.
In 2003 \cite{CiVi03} it was realized that the method
could be used for hierarchical clustering of 
natural data sets from arbitrary (also heterogenous) domains, and 
the theory related to the application of real-world compressors
was developed, and numerous applications in different domains
were given, Section~\ref{sect.applncd}. 
In \cite{Ke04} the authors use a simplified version of the similarity
metric, which also performs well.
In \cite{BCL02a}, and follow-up work,
a closely related notion of compression-based
distances is proposed. There
the purpose was initially to infer a language tree 
from different-language text corpora,
as well as do
authorship attribution on basis of text corpora.
The distances determined between objects are
justified by ad-hoc plausibility arguments 
and represent a partially independent
development (although they refer to the 
information distance approach of \cite{LiVi97,BGLVZ}).
Altogether, it appears that the notion
of compression-based similarity metric is so powerful that its
performance is robust under considerable variations.

\section{Similarity Distance}

We briefly outline an improved version of 
the main theoretical contents
of \cite{CiVi03} and its relation to \cite{LCLMV03}. 
For details and proofs see these references.
First, we give a precise formal meaning to the loose
distance notion of ``degree of similarity''
used in the pattern recognition
literature.

\subsection{Distance and Metric}
Let $\Omega$ be a nonempty set and ${\cal R}^+$ be the set of nonnegative
real numbers.
A {\em distance function} on $\Omega$ is a function
$D: \Omega \times \Omega \rightarrow {\cal R}^+$. It is a {\em metric}
if it satisfies the metric (in)equalities:
\begin{itemize}
\item
$D(x,y)=0$ iff $x=y$,
\item
$D(x,y)=D(y,x)$ (symmetry), and
\item
$D(x,y)\leq D(x,z)+D(z,y)$ (triangle inequality).
\end{itemize}
The value $D(x,y)$ is called the {\em distance} between $x,y \in \Omega$.
A familiar example of a distance that is also metric is the Euclidean metric,
the everyday distance $e(a,b)$ between two geographical objects $a,b$
expressed in, say, meters.
Clearly, this distance satisfies the properties
$e(a,a)=0$, $e(a,b)=e(b,a)$, and $e(a,b) \leq e(a,c) + e(c,b)$
(for instance, $a=$ Amsterdam, $b=$ Brussels, and $c=$ Chicago.)
We are interested in a particular type of distance,
the ``similarity distance'', which we formally define
in Definition~\ref{eq.defsm}.
For example, if the objects are classical music pieces
then the function $D$ defined by $D(a,b)= 0$ if $a$ and $b$ are by the same composer
and $D(a,b) = 1$ otherwise,
is a similarity distance that is also a metric.
This metric captures only one similarity aspect (feature)
of music pieces, presumably an important one that subsumes
a conglomerate of more elementary features.

\subsection{Admissible Distance}
In defining a class of admissible distances (not necessarily metric
distances)
we want to exclude unrealistic ones
like $f(x,y) = \frac{1}{2}$ for {\em every} pair $x  \neq y$.
We do this by restricting
the number of objects within a given distance of an object.
As in \cite{BGLVZ} we do this by only considering effective distances,
as follows.

\begin{definition}\label{def.em}
\rm
Let $\Omega = \Sigma^*$, with $\Sigma$ a finite nonempty alphabet
and $\Sigma^*$ the set of finite strings over that alphabet.
Since every finite alphabet can be recoded in binary,
we choose $\Sigma = \{0,1\}$. In particular,
``files'' in computer memory are finite binary strings.
A function $D: \Omega \times \Omega \rightarrow {\cal R}^+$ is an
{\em admissible distance} if for every pair of objects $x,y \in \Omega$
the distance  $D(x,y)$ satisfies the {\em density} condition
\begin{equation}\label{eq.kraft}
\sum_y 2^{-D(x,y)} \leq 1,
\end{equation}
is {\em computable}, and is {\em symmetric}, $D(x,y)=D(y,x)$.
\end{definition}

If $D$ is an admissible distance, then for every $x$ the
set $\{D(x,y): y \in \{0,1\}^*\}$ is the length set of a prefix code,
since it satisfies \eqref{eq.kraft},
the Kraft inequality. Conversely, if a distance is the length set
of a prefix code, then it satisfies \eqref{eq.kraft}, 
see for example \cite{LiVi97}.

\subsection{Normalized Admissible Distance}
Large objects (in the sense of long strings)
that differ by a tiny part are intuitively
closer than tiny objects that differ by the same amount.
For example, two whole mitochondrial genomes
of 18,000 bases that differ by 9,000 are very different, while two whole
nuclear genomes of $3 \times 10^9$ bases
 that differ by only 9,000 bases are very similar.
Thus, absolute difference between two objects doesn't govern similarity,
but relative difference appears to do so.
\begin{definition}
\rm
A {\em compressor} is a lossless encoder mapping $\Omega$ into
$\{0,1\}^*$ such that the resulting code is a prefix code.
``Lossless'' means that there is a decompressor that reconstructs
the source message from the code message.
For convenience of notation we identify ``compressor''
with a ``code word length function'' $C: \Omega \rightarrow {\cal N}$,
where ${\cal N}$ is the set of nonnegative integers. That is,
the compressed
version of a file $x$ has length $C(x)$.
We only consider compressors such that $C(x) \leq |x|+O(\log |x|)$.
(The additive logarithmic term is due to our requirement
that the compressed file be a prefix code word.)
We fix a compressor $C$,
and call the fixed compressor
the {\em reference compressor}.
\end{definition}
\begin{definition}
\rm
Let $D$ be an admissible distance. Then
$D^+(x)$ is defined by $D^+(x)=\max \{ D(x,z): C(z) \leq C(x)\}$,
and $D^+(x,y)$ is defined by
$D^+(x,y) = \max \{ D^+(x),D^+(y) \}$.
Note that since $D(x,y)=D(y,x)$,
also $D^+(x,y)=D^+(y,x)$.
\end{definition}

\begin{definition}\label{eq.defsm}
\rm
Let $D$ be an admissible distance. The
{\em normalized admissible distance},
also called a {\em similarity distance},  $d(x,y)$,
based on $D$ relative to a reference compressor $C$,
is defined by
\[
d(x,y) = \frac{D(x,y)}{D^+(x,y)}.
\]
\end{definition}

It follows from the definitions that
a normalized admissible distance is a function
$d: \Omega \times \Omega \rightarrow [0,1]$
that is symmetric: $d(x,y)=d(y,x)$.

\begin{lemma}
For every $x \in \Omega$, and constant $e \in [0,1]$,
 a normalized admissible distance satisfies the density constraint
\begin{equation}\label{eq.dp}
 | \{y: d(x,y) \leq e, \;\; C(y) \leq C(x)  \} | < 2^{e  D^+(x)+1} .
\end{equation}
\end{lemma}

We call a normalized distance a ``similarity'' distance, because
it gives a relative similarity according to the distance
(with distance 0 when objects are maximally similar and distance 1 when
they are maximally dissimilar)
and, conversely, for every well-defined computable notion of similarity we
can express it as a metric distance according to our definition.
In the literature a distance that expresses lack of similarity (like
ours)  is often
called a ``dissimilarity'' distance or a ``disparity'' distance.

\subsection{Normal Compressor}
\label{sect.nc}
We give axioms determining a large 
family of compressors that both include most
(if not all) real-world
compressors and ensure the desired
properties of the \NCD to be defined later.

\begin{definition}
\rm
A compressor $C$ is {\em normal}
if it satisfies, up to an additive $O(\log n)$ term,
with $n$ the maximal binary length of an element of $\Omega$
involved in the (in)equality concerned,
the following:
\begin{enumerate}
\item
{\em Idempotency}: $C(xx)=C(x)$, and $C(\lambda)=0$, where
$\lambda$ is the empty string.
\item
{\em Monotonicity}: $C(xy) \geq C(x)$.
\item
{\em Symmetry}: $C(xy)=C(yx)$.
\item
{\em Distributivity}:
$C(xy) + C(z) \leq C(xz)+C(yz)$.
\end{enumerate}
\end{definition}
\begin{remark}
\rm
These axioms are of course an idealization. The reader can 
insert, say $O(\sqrt{n})$, for the $O(\log n)$ fudge term,
and modify the subsequent discussion accordingly. Many
compressors, like gzip or bzip2, have a bounded window size.
Since compression of objects exceeding the window size is
not meaningful, we assume $2n$ is less than the window size.
In such cases the $O(\log n)$ term, or its equivalent, relates
to the fictitious version of the compressor where the
window size can grow indefinitely. Alternatively, we bound the
value of $n$ to half te window size, and replace the fudge term $O(\log n)$ by
some small fraction of $n$.  Other compressors,
like PPMZ, have unlimited window size, and hence are more suitable
for direct interpretation of the axioms.
\end{remark}

{\bf Idempotency:}
A reasonable compressor will see exact repetitions and obey
idempotency up to the required precision.
It will also compress the empty string to the empty string.

{\bf Monotonicity:}
A real compressor
must have the monotonicity property, at least up to the required
precision. The property is evident for stream-based compressors,
and only slightly less evident for block-coding compressors.

{\bf Symmetry:}
Stream-based compressors of the Lempel-Ziv family, like gzip and pkzip, and
the predictive PPM family, like PPMZ, are possibly not precisely
symmetric.
This is related to the stream-based property: the initial file $x$
may have regularities to which the compressor adapts;
after crossing the border to $y$ it must unlearn those regularities
and adapt to the ones of $x$. This process may cause some imprecision
in symmetry that vanishes asymptotically with the length of $x,y$.
A compressor must be poor indeed (and will certainly
not be used to any extent) if it doesn't satisfy symmetry up
to the required precision.
Apart from stream-based, the other major family of compressors
is block-coding based, like bzip2.
They essentially analyze the
full input block by considering all rotations in obtaining
the compressed version. It is to a great
extent symmetrical, and real  experiments show no
departure from symmetry.

{\bf Distributivity:}
The distributivity property is not immediately intuitive.
In Kolmogorov complexity theory the stronger distributivity
property
\begin{equation}\label{eq.sdistr}
C(xyz)+C(z) \leq C(xz)+C(yz)
\end{equation}
holds (with $K=C$). However, to prove
the desired properties of \NCD below, only the weaker
distributivity property
\begin{equation}\label{eq.wdistr}
C(xy)+C(z) \leq C(xz)+C(yz)
\end{equation}
above is required,
also for the boundary case were $C=K$. In practice, real-world
compressors appear to satisfy this weaker distributivity property up to
the required precision.
\begin{definition}
\rm
Define
\begin{equation}\label{eq.cci}
C(y|x)=C(xy)-C(x).
\end{equation}
This number $C(y|x)$ of bits of information in $y$, relative to $x$,
can be viewed as the excess number of bits in
the compressed version of $xy$ compared to the compressed
version of $x$,  and is called
the amount of {\em conditional compressed information}.
\end{definition}
In the definition of compressor the decompression algorithm is
not included (unlike the case of Kolmorogov complexity, where
the decompressing algorithm is given by definition), but
it is easy to construct one:
Given the compressed version of $x$ in $C(x)$ bits,
we can run the compressor on
all candidate strings $z$---for example, in length-increasing lexicographical
order, until we find the compressed string $z_0=x$. Since this
string decompresses to $x$ we have found $x=z_0$.
Given the compressed version of $xy$ in $C(xy)$ bits,
we repeat this process using strings $xz$ until
we find the string $xz_1$ of which the compressed version
equals the compressed version of $xy$. Since
the former compressed version decompresses to $xy$, we have found $y=z_1$.
By the unique decompression property we find that $C(y|x)$ is the
extra number of bits we require to describe $y$ apart from
describing $x$.
It is intuitively acceptable that the conditional compressed information
$C(x|y)$ satisfies the triangle inequality
\begin{equation}\label{eq.sap}
C(x|y) \leq C(x|z)+C(z|y).
\end{equation}

\begin{lemma}
Both \eqref{eq.sdistr} and \eqref{eq.sap} imply \eqref{eq.wdistr}.
\end{lemma}

\begin{lemma}
A normal compressor satisfies additionally 
{\em subadditivity}:
$C(xy) \leq C(x)+C(y)$.
\end{lemma}

{\bf Subadditivity:}
The subadditivity property is clearly also required
for every viable compressor, since a compressor may use information
acquired from $x$ to compress $y$. Minor imprecision may arise from
the unlearning effect of
crossing the border between $x$ and $y$,
mentioned in relation to symmetry,
but again this must vanish asymptotically with increasing length of $x,y$.

\subsection{Normalized Information Distance}
\label{sect.kc}
Technically, the {\em  Kolmogorov complexity} of $x$ given $y$ is the length
of the shortest binary program, for the reference universal
prefix Turing machine, that on input $y$ outputs $x$;
it is denoted as $K(x|y)$. For precise definitions, theory and applications,
see \cite{LiVi97}. The Kolmogorov complexity of $x$ is the length
of the shortest binary program with no input that outputs $x$;
it is denoted as $K(x)=K(x|\lambda)$
where $\lambda$ denotes the empty input.
Essentially, the Kolmogorov complexity of a file
is the length of the ultimate compressed version of the file.
In  \cite{BGLVZ} the {\em information distance} $E(x,y)$ was introduced,
defined as the length of the shortest binary program
for the reference universal prefix Turing machine that, with input $x$
computes $y$, and with input $y$ computes $x$. It was shown there that,
up to an additive logarithmic term, $E(x,y) = \max \{K(x|y),K(y|x)\}$.
It was shown also that $E(x,y)$ is a metric, up to negligible violations
of the metric inequalties.  Moreover, it is
universal in the sense that for every admissible distance $D(x,y)$ as in
Definition~\ref{def.em}, $E(x,y) \leq D(x,y)$ up to an additive
constant depending on $D$ but not on $x$ and $y$.
In \cite{LCLMV03},
the normalized version of $E(x,y)$,
 called the {\em normalized information distance}, is defined as
\begin{equation}\label{eq.nid}
\NID (x,y) = \frac{\max\{K(x|y),K(y|x)\}}{\max\{K(x),K(y)\}}.
\end{equation}
It too is a metric, and it is
universal in the sense that this single metric minorizes up to an
negligible additive error term all normalized admissible distances
in the class considered in \cite{LCLMV03}.
Thus, if two files (of whatever type)
are similar (that is, close) according
to the particular feature described by
a particular normalized
 admissible distance (not necessarily metric),
then they are also similar (that is, close)
in the sense of the normalized information metric. This justifies
calling the latter {\em the\/} similarity metric.
We stress once more
that different pairs of objects may have different dominating
features.
Yet every such dominant similarity is detected by the \NID.
However, this metric is based on the
notion of Kolmogorov complexity.
Unfortunately, the Kolmogorov complexity
is non-computable in the
Turing sense.
Approximation
of the denominator of \eqref{eq.nid} by a given compressor $C$
 is straightforward: it is
$\max\{C(x),C(y)\}$. The numerator is more tricky. It can
be rewritten as
\begin{equation}\label{eq.nom}
\max \{ K(x,y)-K(x),  K(x,y)-K(y) \},
\end{equation}
 within logarithmic
additive precision, by the additive property
of Kolmogorov complexity \cite{LiVi97}.
The term $K(x,y)$ represents the length of the shortest
program for the pair $(x,y)$. In compression practice it is easier
to deal with the concatenation $xy$ or $yx$. Again, within logarithmic
precision $K(x,y)=K(xy)=K(yx)$.
Following a suggestion by
Steven de Rooij, one can approximate \eqref{eq.nom} best by
$\min\{C(xy),C(yx)\} - \min \{C(x),C(y)\}$.
Here, and in the later experiments using the CompLearn Toolkit \cite{Ci03},
we simply use $C(xy)$ rather than
$\min\{C(xy),C(yx)\}$. This is justified by the observation
that block-coding based compressors are symmetric
almost by definition, and experiments with various stream-based
compressors (gzip, PPMZ) show only small
deviations from symmetry.

The result of approximating the \NID using a real
compressor $C$ is called the normalized compression distance (\NCD),
formally defined in \eqref{eq.ncd}.
The theory as developed for the Kolmogorov-complexity
based \NID
in \cite{LCLMV03},
may not hold for
the (possibly poorly) approximating \NCD.
It is nonetheless the case that experiments show that the \NCD
apparently has (some) properties that make the \NID so appealing.
To fill this gap between theory and practice,
we develop the theory of \NCD from first principles,
based on the axiomatics
of Section~\ref{sect.nc}.
 We show that the \NCD is a quasi-universal
similarity metric relative to a normal reference compressor $C$.
The theory developed in \cite{LCLMV03} is
the boundary case $C=K$, where the ``quasi-universality''
below has become full ``universality''.

\subsection{Compression Distance}
We define a compression distance based on a normal compressor
and show it is an admissible distance.
In applying
the approach, we have to make do with an approximation based on a
far less powerful real-world reference compressor $C$.
A compressor $C$ approximates the information distance $E(x,y)$,
 based on Kolmogorov complexity,
by the compression distance $E_C(x,y)$ defined as
\begin{equation}
E_C(x,y) = C(xy)- \min \{C(x),C(y)\}.
\end{equation}
Here,
$C(xy)$ denotes the compressed size of the concatenation of $x$ and $y$,
$C(x)$ denotes the compressed size of $x$,
and $C(y)$ denotes the compressed size of $y$.

\begin{lemma}\label{lem.ncad}
If $C$ is a normal compressor, then
$E_C (x,y)+O(1)$ is an admissible distance.
\end{lemma}

\begin{lemma}
If $C$ is a normal compressor, then
$E_C(x,y)$ satisfies the metric
(in)equalities up to logarithmic additive precision.
\end{lemma}

\begin{lemma}\label{lem.nc+}
If $C$ is a normal compressor, then
$E^+_C (x,y)= \max \{C(x), C(y) \}$.
\end{lemma}

\subsection{Normalized Compression Distance}
The normalized version of the admissible distance
$E_C(x,y)$, the compressor $C$ based approximation of the normalized
information distance  \eqref{eq.nid}, is called the
{\em normalized compression distance} or $\NCD$:
\begin{equation}\label{eq.ncd}
\NCD(x,y) = \frac{C(xy)- \min \{C(x),C(y)\}}{ \max\{C(x),C(y)\}}.
\end{equation}
This $\NCD$
is the main concept of this work. It is the real-world
version of the ideal notion of normalized information distance
\NID in \eqref{eq.nid}.
Actually, the
\NCD is a family of compression functions parameterized 
by the given data
compressor $C$. 

\begin{remark}
\rm
In practice,
the \NCD is a non-negative number
$0 \leq  r \leq 1 + \epsilon$ representing how
different the two files are. Smaller numbers represent more similar files.
The $\epsilon$ in the upper bound is due to
imperfections in our compression techniques,
but for most standard compression algorithms one is unlikely
to see an $\epsilon$ above 0.1 (in our experiments gzip and bzip2 achieved
\NCD's above 1, but PPMZ always had \NCD at most 1).
\end{remark}

There is a natural interpretation to $\NCD(x,y)$: If, say, $C(y) \geq C(x)$
then we can rewrite
\[\NCD(x,y) = \frac{C(xy)-C(x)}{C(y)} . \]
That is, the distance $\NCD(x,y)$ between $x$ and $y$ is the
improvement due to compressing $y$ using $x$ as previously compressed
``data base,'' and compressing $y$ from scratch,
expressed as the ratio between the bit-wise length of the two
compressed versions.
Relative to the reference compressor we
can define the information  in $x$ about $y$ as $C(y)-C(y|x)$. Then,
using \eqref{eq.cci},
\[
\NCD(x,y) = 1 - \frac{C(y)-C(y|x)}{C(y)}.
\]
That is, the \NCD between $x$ and $y$ is 1 minus the ratio of the
information $x$ about $y$ and the information in $y$.

\begin{theorem}
If the compressor is normal, then the \NCD is a normalized admissible
distance satsifying the metric (in)equalities, that is, a similarity metric.
\end{theorem}

{\bf Quasi-Universality:}
We now digress to the theory developed in \cite{LCLMV03}, which
formed the motivation for developing the \NCD.
If, instead of the result of some real compressor,
we substitute the Kolmogorov
complexity for the lengths of the compressed files in the \NCD
formula, the result is the \NID as in \eqref{eq.nid}.
It is universal
in the following sense: Every 
admissible distance expressing similarity
according to some feature,
that can be computed from the objects concerned, is comprised
(in the sense of minorized) by the \NID.
Note that every feature of the data gives rise to a similarity,
and, conversely, every similarity can be thought of
as expressing some feature:
being similar in that sense.
Our actual practice in using the \NCD falls short of
this ideal theory in at least
three respects:

(i) The claimed universality of the \NID
holds only for indefinitely long sequences $x,y$. Once we consider
strings $x,y$ of definite length $n$, it
is only universal with respect to ``simple'' computable
normalized admissible
distances, where ``simple'' means that they are computable by programs
of length, say, logarithmic in $n$.
This reflects the fact that, technically speaking, the universality
is achieved by summing the weighted contribution of all
similarity distances in the class considered with respect
to the objects considered. Only similarity distances of which
the complexity is small (which means that the weight is large),
with respect to the size of the data concerned, kick in.

(ii) The Kolmogorov complexity is not computable, and it is
in principle impossible to compute how far off the
\NCD is from the \NID. So we cannot in general know
how well we are doing using the \NCD of a given compressor.
Rather than all ``simple'' distances (features, properties), 
like the \NID, the \NCD captures a subset of these based on the features
(or combination of features) analyzed by the compressor. For natural
data sets, however, these may well cover the features and regularities
present in the data anyway. Complex features, expressions of simple
or intricate computations, like the initial segment of $\pi=3.1415 \ldots$,
seem unlikely to be hidden in natural data. This fact may account for
the practical success of the \NCD, especially when using good compressors.

(iii) To approximate the \NCD
we use standard compression programs like gzip, PPMZ, and bzip2.
While better compression
of a string will always  approximate the Kolmogorov complexity better,
this may not be true for the \NCD. Due to its arithmetic
form, subtraction and division, it is theoretically possible
that while all items in the formula get better compressed,
the improvement is not the same for all items, and the \NCD value
moves away from the \NID value.
In our experiments we have not observed this behavior in a noticable fashion.
Formally, we can state the following:

\begin{theorem}\label{theo.uni}
Let $d$ be a computable normalized admissible distance and $C$
be a normal compressor.
Then, $\NCD(x,y) \leq \alpha d(x,y) + \epsilon$, where
for $C(x) \geq C(y)$, we have
$\alpha = D^+(x)/ C(x)$
 and $\epsilon = (C(x|y)-K(x|y))/C(x)$,
with $C(x|y)$ according to \eqref{eq.cci}.
\end{theorem}

\begin{remark}
\rm
Clustering according to \NCD will group sequences together that
are similar according to features that are not
explicitly known to us. Analysis
of what the compressor actually does, still  may not tell us which
features that make sense to us can be expressed by conglomerates
of features analyzed by the compressor. This can be exploited to
track down unknown features implicitly in classification: forming automatically
clusters of data and see in which cluster (if any) a new candidate
is placed.

Another aspect that can be exploited is exploratory:
Given that the \NCD is small for a pair $x,y$ of specific sequences,
what does this really say about the sense in which these two sequences are
similar?
The above analysis suggests that close
similarity will be due to a dominating feature (that perhaps expresses
a conglomerate of subfeatures). Looking into these deeper causes may give
feedback about the appropriateness of the realized \NCD distances and may help
extract more intrinsic information about the objects,
than the oblivious division into clusters, by
looking for the common features in the data clusters.
\end{remark}

\subsection{Hierarchical Clustering}
Given a set of objects, the pairwise \NCD's form
the entries of a distance matrix.
This distance matrix contains the pairwise relations
in raw form. But in this format
that information is not easily usable.
Just as the distance matrix is a reduced form of information
representing the original data set, we now need to reduce the
information even further in order to achieve a cognitively acceptable
format like data clusters.
The distance matrix contains all the information in a form that 
is not easily usable, since for $n > 3$ our cognitive capabilities 
rapidly fail.
In our situation we do not know the number of clusters a-priori, 
and we let the data decide the clusters. 
The most natural way to do so is hierarchical clustering \cite{DHS}. 
Such methods have been extensively investigated
in Computational Biology 
in the context of producing phylogenies of species. 
One the most sensitive ways
is the so-called `quartet method.’ This method is sensitive, 
but time consuming, running in quartic time. 
Other hierarchical clustering methods, like parsimony, may be 
much faster, quadratic time, but they are less sensitive. 
In view of the fact that current compressors are good but limited, 
we want to exploit the smallest differences in distances,
 and therefore use the most sensitive method to get greatest 
accuracy. Here, we use a new quartet-method (actually a 
new version \cite{CiVi03} of the quartet puzzling variant 
\cite{SvH}), which is a heuristic based on randomized parallel 
hill-climbing genetic programming. In this paper we do not 
describe this method in any detail, 
the reader is referred to \cite{CiVi03},
or the full description in \cite{CiVi05}.
It is implemented in the CompLearn package \cite{Ci03}. 

We describe the idea of the algorithm,
 and the interpretation of the accuracy of 
the resulting tree representation of the data clustering. 
To cluster $n$ data items, the algorithm generates a 
random ternary tree with $n-2$ internal nodes and $n$ leaves.  
The algorithm tries to improve the solution at each step 
by interchanging sub-trees rooted at internal nodes (possibly leaves).
 It switches if the total tree cost is improved. 
To find the optimal tree is NP-hard, that is, 
it is infeasible in general.  
To avoid getting stuck in a local optimum,
 the method executes sequences of elementary mutations 
in a single step. The length of the sequence is drawn from a fat tail 
distribution, to ensure that the probability of drawing a 
longer sequence is still significant.  
In contrast to other methods, this guarantees that, 
at least theoretically, in the long run a global optimum is achieved. 
Because the problem is NP-hard, we can not expect the 
global optimum to be reached in a feasible time in general. 
Yet for natural data, like in this work, experience shows that 
the method usually reaches an apparently global optimum. 
One way to make this more likely is to run several optimization 
computations in parallel, and terminate only when they all agree 
on the solutions (the probability that this would arises by 
chance is very low as for a similar technique in Markov chains). 
The method is so much improved against previous 
quartet-tree methods, that it can cluster larger groups 
of objects (around 70) than was previously possible (around 15). 
If the latter methods need to cluster groups larger than 15, 
they first cluster sub-groups into small trees 
and then combine these trees by a super-tree reconstruction method. 
This has the drawback that optimizing the 
local subtrees determines relations that 
cannot be undone in the supertree construction, 
and it is almost guaranteed that such methods cannot 
reach a global optimum.
Our clustering heuristic generates a tree with a certain fidelity
with respect to the underlying distance matrix (or alternative data
from which the quartet tree is constructed)
called standardized benefit score or $S(T)$ value in the sequel. 
This value measures the quality  of the tree 
representation of the overall 
oder relations between the distances 
in the matrix. It measures in how far the tree can represent
the quantitative distance relations in a topological qualitative
manner without violating relative order. The $S(T)$ value ranges from 0 (worst) to 
1 (best). A random tree is likely to have $S(T) \approx 1/3$, 
while $S(T)=1$  means that the relations in the distance matrix 
are perfectly represented by the tree. Since we
deal with $n$ natural data objects, living in a space
of unknown metric,
we know a priori only that the pairwise distances between them
can be truthfully represented in $n-1$-dimensional Euclidian space. 
Multidimensional scaling, representing the data by points 
in 2-dimensional space, most likely necessarily distorts 
the pairwise distances. This is akin to the distortion arising 
when we map spherical earth geography on a flat map. 
A similar thing happens if we represent the $n$-dimensional 
distance matrix by a ternary tree. It can be shown that some
5-dimensional distance matrices can only be mapped in a ternary 
tree with $S(T) < 0.8$. Practice shows, however, that up to 
12-dimensional distance matrices, arising from natural data, 
can be mapped into a such tree with very little distortion 
($S(T)>0.95$). In general the $S(T)$ value deteriorates for 
large sets. The reason is that, with increasing size of natural 
data set, the projection of the information in the distance matrix 
into a ternary tree gets necessarily increasingly distorted. 
If for a large data set like 30 objects, the $S(T)$ value is large, 
say $S(T)\geq 0.95$, then this gives evidence that the tree 
faithfully represents the distance matrix, but also that 
the natural relations between this large set of data 
were such that they could be represented by such a tree.

\section{Applications of NCD}
\label{sect.applncd}
The compression-based \NCD method to
establish a universal similarity metric \eqref{eq.ncd} among objects
given as finite binary strings,
and, apart from what was mentioned in the Introduction,
has been applied to
objects like music pieces in MIDI format, \cite{CWV03},
computer programs,
genomics, virology, language tree of non-indo-european
languages, literature in Russian 
Cyrillic and English translation, optical character recognition
of handwrittern digits in simple bitmap formats, 
or astronimical time sequences, and combinations 
of objects from heterogenous
domains, using statistical, dictionary, and block sorting 
compressors,
\cite{CiVi03}. In \cite{Ke04}, the authors compared the
performance of the method on all major time sequence data bases
used in all major data-mining conferences in the last decade,
against all major methods. It turned out that the \NCD method 
was far superior to any other method
in  heterogenous data clustering and anomaly detection and
performed comparable to the other methods in the simpler tasks.
We developed the CompLearn Toolkit, \cite{Ci03}, and performed
experiments in vastly different
application fields to test the quality and universality of the method.
In \cite{We04}, the method is used to analyze network traffic 
and cluster computer worms and virusses. Currently, a plethora
of new applications of the method arise around the world, in many
areas, as the reader can verify by searching for the
papers `the similarity metric'
or `clustering by compression,' and look at the papers
that refer to these, in Google Scholar.
\begin{figure}[htb]
\begin{center}
\epsfig{file=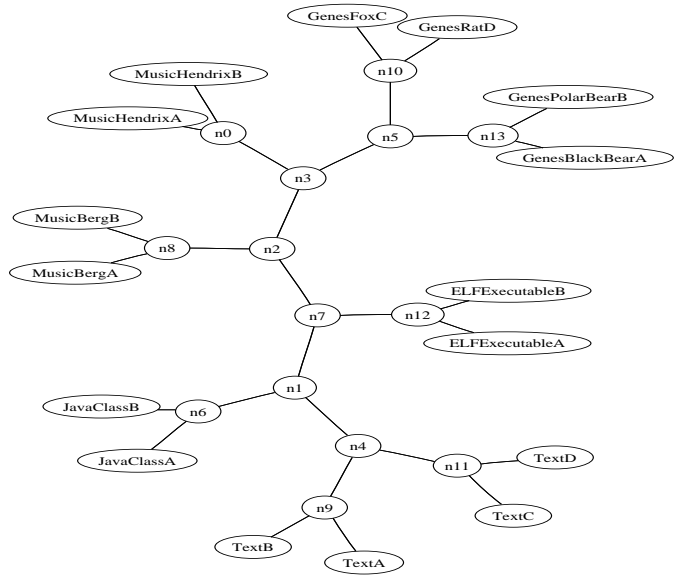,height=3in,width=3.5in}
\end{center}
\caption{Classification of different file types.
Tree agrees
exceptionally well with
\NCD distance matrix: $S(T)=0.984$.}\label{figfiletypes}
\end{figure}
\subsection{Heterogenous Natural Data}

The success of the method as reported depends strongly on the
judicious use of encoding of the objects compared. Here one should
use common sense on what a real world compressor can do. There are
situations where our approach fails if applied in a
straightforward way.
For example: comparing text files by the same authors
in different encodings (say, Unicode and 8-bit version) 
is bound to fail.
For the ideal similarity metric  based on
Kolmogorov complexity as defined in \cite{LCLMV03}
this does not matter at all, but for
practical compressors used in the experiments it will be fatal.
Similarly, in the music experiments we use symbolic MIDI
music file  format rather than wave-forms.
We test gross classification of files
based on heterogenous data of markedly different file types:
(i) Four mitochondrial gene sequences, from a black bear, polar bear,
fox, and rat obtained from the GenBank Database on the world-wide web;
(ii) Four excerpts from the novel { \em The Zeppelin's Passenger} by
E.~Phillips Oppenheim, obtained from the Project Gutenberg Edition on the World-Wide web;
(iii) Four MIDI files without further processing; two from Jimi Hendrix and
two movements from Debussy's Suite Bergamasque, downloaded from various
repositories on the
world-wide web;
(iv) Two Linux x86 ELF executables (the {\em cp} and {\em rm} commands),
copied directly from the RedHat 9.0 Linux distribution; and
(v)  Two compiled Java class files, generated by ourselves.
The compressor used to compute the \NCD matrix was bzip2.
As expected, the program correctly classifies each of the different types
of files together with like near like. The result is reported
in Figure~\ref{figfiletypes} with $S(T)$ equal to the very high
confidence value 0.984.
This experiment shows the power and universality of the method:
no features of any specific domain of application are used.
We believe that there is no other method known that can cluster
data that is so heterogenous this reliably. This is borne out by the
massive experiments with the method in \cite{Ke04}.

\begin{figure}
\hfill\ \psfig{figure=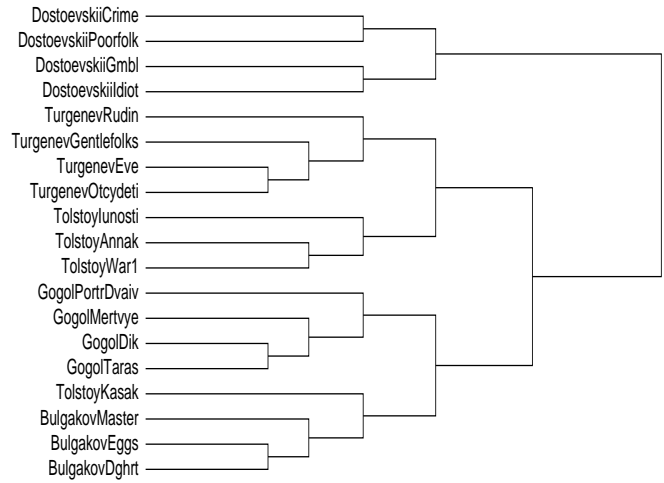,width=3.5in,height=2.5in} \hfill\
\caption{Clustering of Russian writers.
Legend: I.S. Turgenev, 1818--1883 [Father and Sons, Rudin, On the Eve,
A House of Gentlefolk]; F. Dostoyevsky 1821--1881 [Crime and Punishment,
The Gambler,
The Idiot; Poor Folk]; L.N. Tolstoy 1828--1910 [Anna Karenina, The Cossacks,
Youth, War and Piece]; N.V. Gogol 1809--1852 [Dead Souls, Taras Bulba,
The Mysterious Portrait,  How the Two Ivans Quarrelled];
M. Bulgakov 1891--1940 [The Master and Margarita, The Fatefull Eggs, The
Heart of a Dog]. $S(T)=0.949$.
}\label{fig.russwriter}
\end{figure}

\subsection{Literature}
The texts used in this experiment were down-loaded from the world-wide web
in original Cyrillic-lettered Russian and in Latin-lettered English
by L. Avanasiev.
\begin{figure}
\begin{center}
\epsfig{file=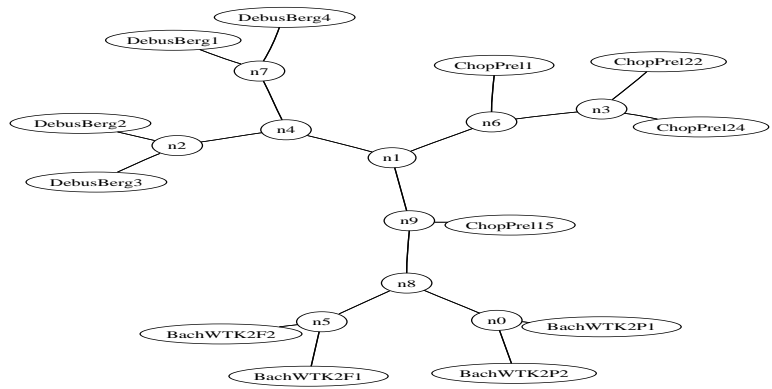,width=4in,height=2in}
\end{center}
\caption{Output for the 12-piece set.
Legend: J.S. Bach [Wohltemperierte Klavier II: Preludes and Fugues 1,2---
BachWTK2\{F,P\}\{1,2\}]; Chopin [Pr\'eludes op.~28: 1, 15, 22, 24
---ChopPrel\{1,15,22,24\}];
Debussy [Suite Bergamasque, 4 movements---DebusBerg\{1,2,3,4\}].
$S(T)=0.968$.}\label{figsmallset}
\end{figure}
The compressor used to compute the \NCD matrix was bzip2.
We clustered Russian literature in the original
(Cyrillic) by Gogol, Dostojevski, Tolstoy, Bulgakov,Tsjechov,
with three or four different texts per author. Our purpose was to
see whether the clustering is sensitive enough, and the authors distinctive
enough, to result in clustering by
author. In Figure~\ref{fig.russwriter} we see an 
almost perfect clustering according to author.
Considering the English translations of the same texts,
we saw errors in the clustering (not shown).
Inspection showed that the clustering was now partially based on the
translator. It appears that the translator superimposes his characteristics
on the texts, partially suppressing 
the characteristics of the original
authors.
In other experiments, not reported here,
 we separated authors by gender and by period.

\subsection{Music}\label{secdetails}
The amount of digitized music available on the internet has grown
dramatically in recent years, both in the public domain
and on commercial sites. Napster and its clones are prime examples.
Websites offering musical content in some form or other
(MP3, MIDI, \ldots) need a way to organize their wealth of material;
they need to somehow classify their files according to
musical genres and subgenres, putting similar pieces together.
The purpose of such organization is to enable users
to navigate to pieces of music they already know and like,
but also to give them advice and recommendations
(``If you like this, you might also like\ldots'').
Currently, such organization is mostly done manually by humans,
but some recent research has been looking into the possibilities
of automating music classification. For details about the music 
experiments see \cite{CWV03,CiVi03}.
\subsection{Bird-Flu Virii---H5N1}
\begin{figure}
\begin{center}
\epsfig{file=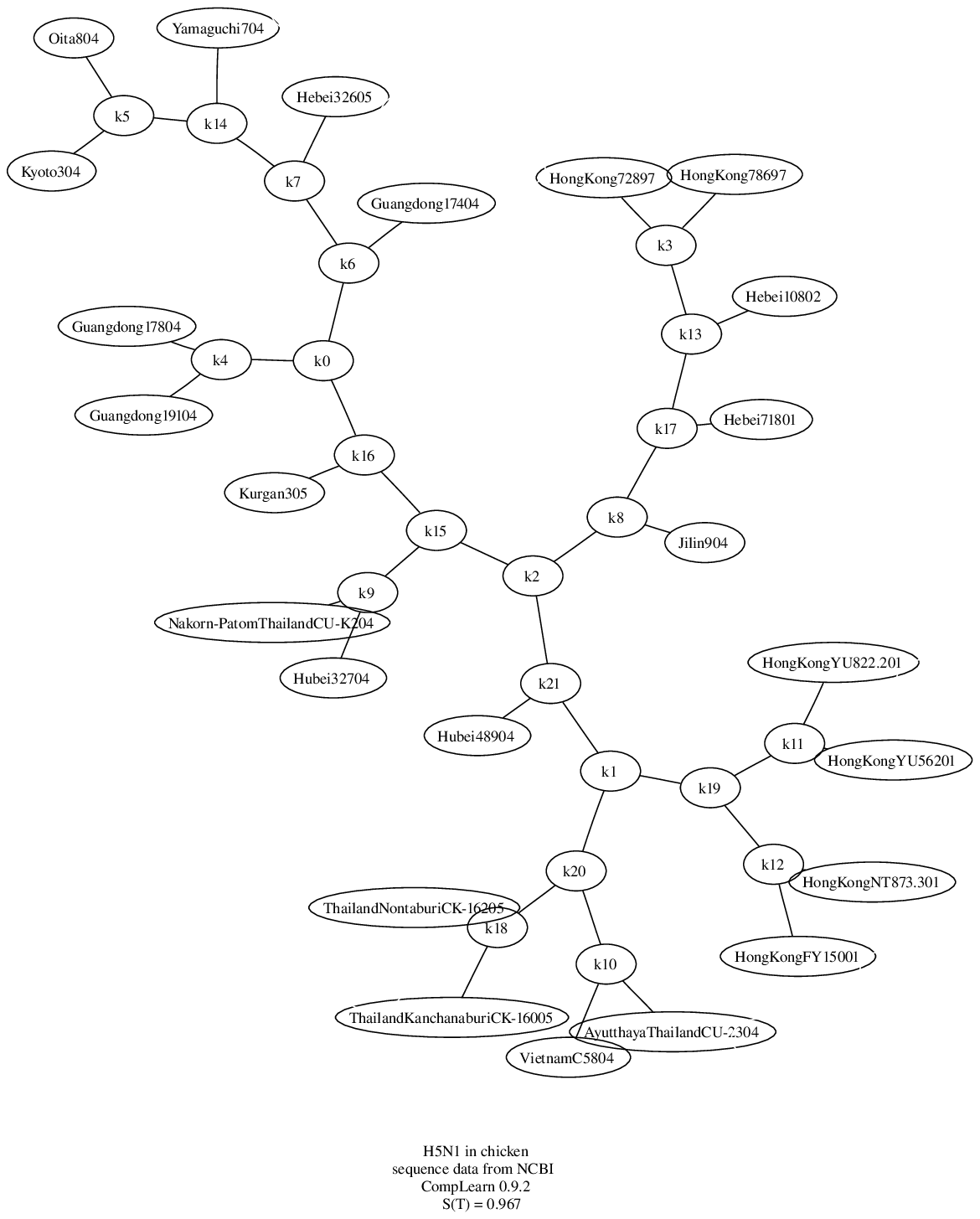,width=4in,height=3in}
\end{center}
\caption{Set of 24 Chicken Examples of H5N1 Virii. 
$S(T)=0.967$.}\label{figbirdflu}
\end{figure}
In Figure~\ref{figbirdflu} we display classification of bird-flu
virii of the type H5N1 that have been found in different geographic 
locations in chicken. Data downloaded from the 
National Center for Biotechnology Information (NCBI),
National Library of Medicine, National Institutes of Health (NIH).

\section{Google-Based Similarity}
To make computers more intelligent one would like
to represent meaning in computer-digestable form.
Long-term and labor-intensive efforts like
the {\em Cyc} project \cite{cyc:intro} and the {\em WordNet}
project \cite{wordnet} try to establish semantic relations
between common objects, or, more precisely, {\em names} for those
objects. The idea is to create
a semantic web of such vast proportions that rudimentary intelligence
and knowledge about the real world spontaneously emerges.
This comes at the great cost of designing structures capable
of manipulating knowledge, and entering high
quality contents in these structures
by knowledgeable human experts. While the efforts are long-running
and large scale, the overall information entered is minute compared
to what is available on the world-wide-web.

The rise of the world-wide-web has enticed millions of users
to type in trillions of characters to create billions of web pages of
on average low quality contents. The sheer mass of the information
available about almost every conceivable topic makes it likely
that extremes will cancel and the majority or average is meaningful
in a low-quality approximate sense. We devise a general
method to tap the amorphous low-grade knowledge available for free
on the world-wide-web, typed in by local users aiming at personal
gratification of diverse objectives, and yet globally achieving
what is effectively the largest semantic electronic database in the world.
Moreover, this database is available for all by using any search engine
that can return aggregate page-count estimates like Google for a large
range of search-queries.

The crucial
point about the \NCD method above is that 
the method analyzes the objects themselves.
This precludes comparison of abstract notions or other objects
that don't lend themselves to direct analysis, like
emotions, colors, Socrates, Plato, Mike Bonanno and Albert Einstein.
While the previous \NCD method that compares the objects themselves using
\eqref{eq.ncd} is
particularly suited to obtain knowledge about the similarity of
objects themselves, irrespective of common beliefs about such
similarities, we now develop a method that uses only the name
of an object and obtains knowledge about the similarity of objects
by tapping available information generated by multitudes of
web users.
The new method is useful to extract knowledge from a given corpus of
knowledge, in this case the Google database, but not to
obtain true facts that are not common knowledge in that database.
For example, common viewpoints on the creation myths in different
religions
may be extracted by the Googling method, but contentious questions
of fact concerning the phylogeny of species can be better approached
by using the genomes of these species, rather than by opinion.

{\bf Googling for Knowledge:}
Let us start with simple intuitive justification (not to be mistaken
for a substitute of the underlying mathematics)
 of the approach we propose in \cite{CV04}.
While the theory we propose is rather intricate, the resulting method
is simple enough. We give an example:
At the time of doing the experiment, a Google search
for ``horse'', returned
46,700,000 hits. The number of hits for the
search term ``rider'' was 12,200,000. Searching
for the pages where both ``horse'' and ``rider'' occur gave
 2,630,000 hits, and
Google indexed 8,058,044,651 web pages.
Using these numbers in the main formula  \eqref{eq.NGD} we derive below, with
$N=8,058,044,651$, this yields a Normalized Google Distance
 between the terms ``horse''
and ``rider'' as follows:
\[
\NGD(horse,rider)
\approx 0.443.
\]
In the sequel of the paper we argue that the \NGD is a normed semantic
distance between the terms in question,
usually in between 0 (identical) and 1 (unrelated), in
the cognitive space invoked by the usage of the terms
on the world-wide-web as filtered by Google. Because of the vastness
and diversity of the web this may be taken as related to the current objective meaning
of the terms in society.
We did the same calculation when Google indexed only one-half
of the current number of pages: 4,285,199,774. It is instructive that the
probabilities of the used search terms didn't change significantly over
this doubling of pages, with number of hits for ``horse''
equal 23,700,000, for ``rider'' equal 6,270,000, and
for ``horse, rider'' equal to 1,180,000.
The $\NGD(horse,rider)$ we computed
in that situation was $\approx 0.460$. This is in line with our contention
that the relative frequencies of web pages containing
search terms gives objective information about the semantic
relations between the search terms. If this is the case, then
the Google probabilities of search terms and the computed \NGD's
should stabilize (become scale invariant) with a growing Google database.

{\bf Related Work:}
There is a great deal of work in both 
cognitive psychology \cite{LD97},
linguistics, and computer science, about using word (phrases)
frequencies in text corpora to develop measures for word similarity or word association, partially surveyed in \cite{TC03,TKS02},
going back to at least
\cite{Le69}. One of the most successful is Latent Semantic Analysis
(LSA) \cite{LD97} that has been applied in various forms in a great
number of applications.
As with LSA, many
other previous approaches of extracting meaning from text documents are based
on text corpora that are many order of 
magnitudes smaller, using complex mathematical techniques
like singular value decomposition and dimensionality reduction,
and that are
in local storage,  and on
assumptions that are more restricted, than what we propose.
In contrast, \cite{Ec04,CS04,BbA05} and the many references cited there,
use the web and Google counts to identify 
lexico-syntactic patterns or other data.
Again, the theory, aim, feature analysis,
 and execution are different from ours, and cannot
meaningfully be compared. Essentially, our method below
automatically extracts meaning relations between arbitrary objects
from the web in a manner
that is feature-free,
up to the search-engine used, and computationally feasible.
This seems to be a new direction altogether.

\subsection{The Google Distribution}
Let the set of singleton {\em Google search terms}
be denoted by ${\cal S}$.
In the sequel we use both singleton
search terms and doubleton search terms $\{\{x,y\}: x,y \in {\cal S} \}$.
Let the set of web pages indexed (possible of being returned)
by Google be $\Omega$. The cardinality of $\Omega$ is denoted
by $M=|\Omega|$, and at the time of this writing
$8\cdot 10^9 \leq M \leq 9 \cdot 10^9$
(and presumably greater by the time of reading this).
Assume that a priori all web pages are equi-probable, with the probability
of being returned by Google being $1/M$.  A subset of $\Omega$
is called an {\em event}. Every {\em  search term} $x$ usable by Google
defines a {\em singleton Google event} ${\bf x} \subseteq \Omega$ of web pages
that contain an occurrence of $x$ and are returned by Google
if we do a search for $x$.
Let $L: \Omega \rightarrow [0,1]$ be the uniform mass probability
function.
The probability of
such an event ${\bf x}$ is $L({\bf x})=|{\bf x}|/M$.
 Similarly, the {\em doubleton Google event} ${\bf x} \bigcap {\bf y}
\subseteq \Omega$ is the set of web pages returned by Google
if we do a search for pages containing both search term $x$ and
search term $y$.
The probability of this event is $L({\bf x} \bigcap {\bf y})
= |{\bf x} \bigcap {\bf y}|/M$.
We can also define the other Boolean combinations: $\neg {\bf x}=
\Omega \backslash {\bf x}$ and ${\bf x} \bigcup {\bf y} =
\neg ( \neg {\bf x} \bigcap \neg {\bf y})$, each such event
having a probability equal to its cardinality divided by $M$.
If ${\bf e}$ is an event obtained from the basic events ${\bf x}, {\bf y},
\ldots$, corresponding to basic search terms $x,y, \ldots$,
by finitely many applications of the Boolean operations,
then the probability $L({\bf e}) = |{\bf e}|/M$.
Google events capture in a particular sense
all background knowledge about the search terms concerned available
(to Google) on the web. The Google event ${\bf x}$, consisting of the set of
all web pages containing one or more occurrences of the search term $x$,
thus embodies, in every possible sense, all direct context
in which $x$ occurs on the web.
\begin{remark}
\rm
It is of course possible that parts of
this direct contextual material link to other web pages in which $x$ does not
occur and thereby supply additional context. In our approach this indirect
context is ignored. Nonetheless, indirect context may be important and
future refinements of the method may take it into account.
\end{remark}
The event ${\bf x}$ consists of all
possible direct knowledge on the web regarding $x$.
Therefore, it is natural
to consider code words for those events
as coding this background knowledge. However,
we cannot use the probability of the events directly to determine
a prefix code, or, rather the underlying information content implied
by the probability.
The reason is that
the events overlap and hence the summed probability exceeds 1.
By the Kraft inequality, see for example \cite{LiVi97}, this prevents a
corresponding set of code-word lengths.
The solution is to normalize:
We use the probability of the Google events to define a probability
mass function over the set $\{\{x,y\}: x,y \in {\cal S}\}$
of  Google search terms, both singleton and doubleton terms. There are
$|{\cal S}|$ singleton terms, and
${ |{\cal S}|  \choose 2}$  doubletons consisting of a pair of non-identical
terms.
Define
\[
 N= \sum_{\{x,y\} \subseteq {\cal S}} |{\bf x} \bigcap
{\bf y}|,
\]
counting each singleton set and each doubleton set (by definition
unordered) once in the summation. Note that this means that
for every pair $\{x,y\} \subseteq {\cal S}$, with $x \neq y$,
the web pages $z \in {\bf x} \bigcap
{\bf y}$
 are counted three times: once in ${\bf x} =  {\bf x} \bigcap
{\bf x}$, once in ${\bf y} =  {\bf y} \bigcap
{\bf y}$, and
once in  ${\bf x} \bigcap
{\bf y}$.
Since every web page that is indexed by Google contains at least
one occurrence of a search term, we have $N \geq M$. On the other hand,
web pages contain on average not more than a certain constant $\alpha$
search terms. Therefore, $N \leq \alpha M$.
Define
\begin{align}\label{eq.gpmf}
g(x) = g(x,x), \; \;
g(x,y) =  L({\bf x} \bigcap {\bf y}) M/N =|{\bf x} \bigcap {\bf y}|/N.
\end{align}
Then, $\sum_{\{x,y\} \subseteq {\cal S}} g(x,y) = 1$.
This $g$-distribution changes over time,
and between different samplings
from the distribution. But let us imagine that $g$ holds
in the sense of an instantaneous snapshot. The real situation
will be an approximation of this.
Given the Google machinery, these are absolute probabilities
which allow us to define the associated prefix code-word lengths (information contents)
 for
both the singletons and the doubletons.
The {\em Google code} $G$
is defined by
\begin{align}\label{eq.gcc}
G(x)= G(x,x), \; \;
G(x,y)= \log 1/g(x,y) .
\end{align}
In contrast to strings $x$ where the complexity $C(x)$ represents
the length of the compressed version of $x$ using compressor $C$, for a search
term $x$ (just the name for an object rather than the object itself),
the Google code of length $G(x)$ represents the shortest expected
prefix-code word length of the associated Google event ${\bf x}$.
The expectation
is taken over the Google distribution $p$.
In this sense we can use the Google distribution as a compressor
for Google ``meaning'' associated with the search terms.
The associated \NCD, now called the
{\em normalized Google distance (\NGD)} is then defined
by \eqref{eq.NGD}, and can be rewritten as the right-hand expression: \begin{equation}\label{eq.NGD}
 \NGD(x,y)=\frac{G(x,y) - \min(G(x),G(y))}{\max(G(x),G(y))}
= \frac{  \max \{\log f(x), \log f(y)\}  - \log f(x,y) }{
\log N - \min\{\log f(x), \log f(y) \}},
\end{equation}
where $f(x)$ denotes the number of pages containing $x$, and $f(x,y)$
denotes the number of pages containing both $x$ and $y$, as reported by Google.
This $\NGD$ is an approximation to the $\NID$ of \eqref{eq.nid}
using the prefix code-word lengths (Google code)
generated by the Google distribution as defining a compressor
approximating the length of the Kolmogorov code, using
the background knowledge on the web as viewed by Google
as conditional information. In practice, use the page counts
returned by Google for the frequencies, and we have to
choose $N$.  From the right-hand side term in \eqref{eq.NGD}
it is apparent that by increasing $N$ we decrease the \NGD , everything gets
closer together, and
by decreasing $N$ we increase the \NGD , everything gets further apart.
Our experiments suggest that every reasonable
($M$ or a value greater than any $f(x)$) value can be used as
normalizing factor  $N$,
and our
results seem  in general insensitive to this choice.  In our software, this
parameter $N$ can be adjusted as appropriate, and we often use $M$ for $N$.

{\bf Universality of NGD:} 
In the full paper \cite{CV04} we analyze the mathematical properties of \NGD,
and  prove the universality of the Google distribution among web author based
distributions, as well as the universality of the \NGD with respect to
the family of the individual web author's \NGD's, that is, their
individual semantics relations, (with high probability)---not included here
for space reasons.

\section{Applications}
\label{sect.exp}

\subsection{Colors and Numbers}
\begin{figure}
\centering
\includegraphics[height=3in,width=4.5in]{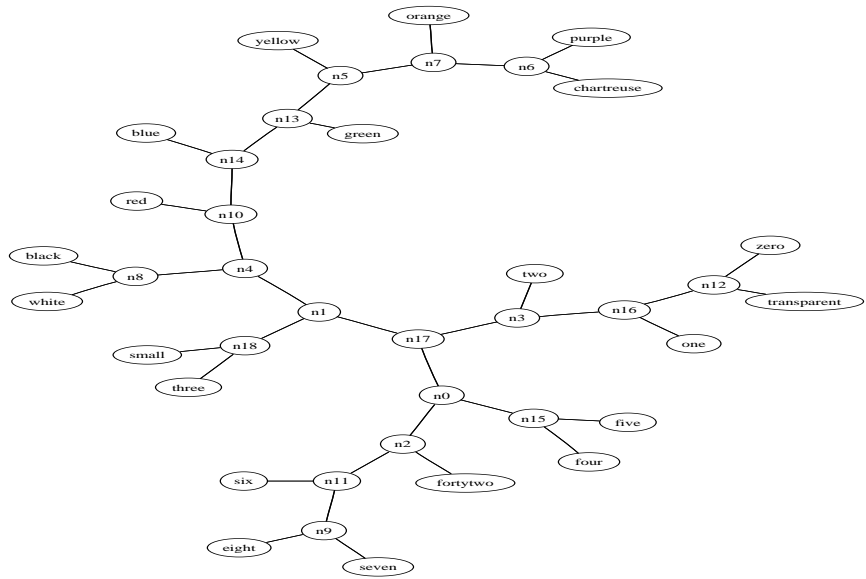}
\caption{Colors and numbers arranged into a tree using \NGD .}
\label{fig.colors}
\end{figure}

The objects to be clustered are search terms
consisting of the names of colors,
numbers, and some tricky words.  The program automatically organized the colors
towards one side of the tree and the numbers towards the other,
Figure~\ref{fig.colors}.
It arranges the terms which have as only meaning a color or a number, and
nothing else, on the farthest reach of the color side
and the number side, respectively. It puts the
more general terms black and white, and zero, one, and two,
towards the center, thus indicating their
more ambiguous interpretation.  Also, things which were not exactly colors
or numbers are also put towards the center, like the word ``small''.
We may consider this an example of automatic ontology creation.
As far as the authors know there do not exist other experiments that
create this type of semantic meaning from nothing (that is, automatically
from the web using Google). Thus, there is no baseline to compare against;
rather the current experiment can be a baseline to evaluate the behavior
of future systems.
\subsection{Names of Literature}
Another example is English novelists. The authors and texts used are:
\begin{figure}
\centering
\includegraphics[height=3in,width=4in]{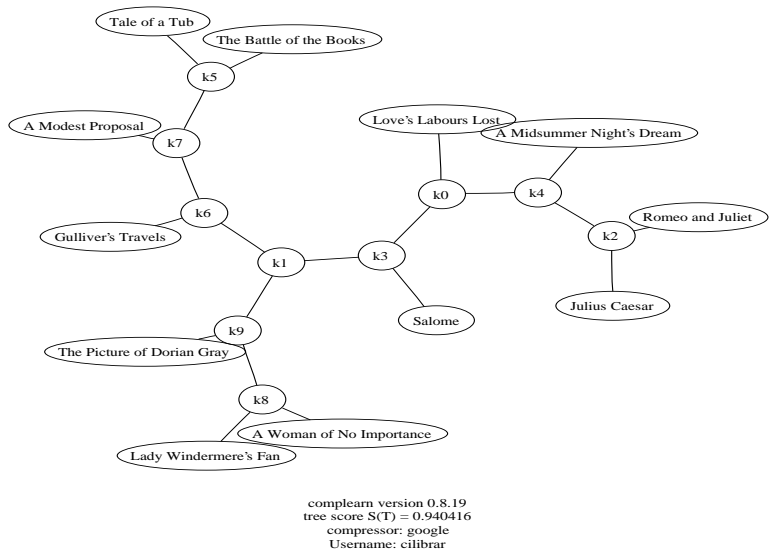}
\caption{Hierarchical clustering of authors. $S(T)=0.940$.}
\label{fig.englishnov}
\end{figure}

\textbf{William Shakespeare:} {\em A Midsummer Night's Dream; Julius Caesar;
Love's Labours Lost; Romeo and Juliet
}.

\textbf{Jonathan Swift:} {\em The Battle of the Books; Gulliver's Travels;
Tale of a Tub; A Modest Proposal};

\textbf{Oscar Wilde:} {\em Lady Windermere's Fan; A Woman of No Importance;
Salome; The Picture of Dorian Gray}.

As search terms we used only the names of texts, without the authors.
The clustering is given in Figure~\ref{fig.englishnov};
it automatically has put the books by the same authors together.
The $S(T)$ value
in Figure~\ref{fig.englishnov}
gives the fidelity of the tree as a representation of the pairwise
distances in the
\NGD matrix (1 is perfect and 0 is as bad as possible. For details
see \cite{Ci03,CiVi03}).
The question arises why we should expect this. Are names of artistic objects
so distinct? (Yes. The point also being that the distances from every single
object to all other objects are involved. The tree takes this global
aspect into account and therefore disambiguates other meanings of the
objects to retain the meaning that is relevant for this collection.)
 Is the distinguishing feature subject matter or title style?
(In these experiments with objects belonging to the cultural
heritage it is clearly a subject matter. To stress the point we
used ``Julius Caesar'' of Shakespeare. This term occurs on the web
overwhelmingly in other contexts and styles. Yet the collection of
the other objects used, and the semantic distance towards those objects,
determined the meaning of ``Julius Caesar'' in this experiment.)
Does the system gets confused if we add more artists? (Representing
the \NGD matrix in bifurcating trees without distortion
 becomes more difficult for, say, more than 25 objects. See \cite{CiVi03}.)
 What about other
subjects, like music, sculpture? (Presumably,
the system will be more trustworthy if the subjects are more common
on the web.) These experiments are representative  for
those we have performed with the current software. For a plethora
of other examples, or to test your own, see the Demo page of \cite{Ci03}.
\subsection{Systematic Comparison with WordNet Semantics}
\label{sect.validation}
WordNet \cite{wordnet} is a semantic concordance
of English.  It focusses on the meaning of words by dividing them
into categories.  We use this as follows. A category we want to learn, the concept,
is termed, say,  ``electrical'', and represents
anything that may pertain to electronics.
The negative examples
are constituted by simply everything else.
This category represents a typical expansion of a node in the
WordNet hierarchy. In an experiment we ran,
the accuracy on the test set is 100\%: It turns
out that ``electrical terms'' are unambiguous and easy to learn
and classify by our method. The information in the WordNet database
is entered over the decades by human experts and is precise. The database
is an academic venture and is publicly accessible. Hence it is
a good baseline against which to judge the accuracy of our method in an indirect
manner. While we cannot directly compare the semantic distance, the \NGD, between
objects, we can indirectly judge how accurate it is by using it as basis
for a learning algorithm. In particular, we investigated how
well semantic categories as learned using the \NGD--\SVM approach
agree with the corresponding WordNet categories.
For details about the structure of WordNet we refer to
the official WordNet documentation available online. We considered 100
randomly selected semantic categories from the WordNet database.
For  each
category we  executed the following sequence.
First, the \SVM is trained on 50 labeled training samples. The positive examples
are randomly drawn from the WordNet database in the category in question.
The negative examples are randomly drwan from a dictionary. While the latter examples
may be false negatives, we consider the probability negligible.
Per experiment we used a total of six anchors,
three of which are randomly drawn from the WordNet database category in question,
and three of which are drawn from the dictionary.
Subsequently, every example is converted to
6-dimensional vectors using \NGD. The $i$th entry of the vector is the
\NGD between the $i$th anchor and the example concerned ($1 \leq i \leq 6$).
 The \SVM is trained on the resulting labeled vectors.
The kernel-width and error-cost parameters are automatically
determined using five-fold cross validation.  Finally, testing of how well
the \SVM has learned the classifier is performed
using 20 new examples in a balanced ensemble of positive and negative
examples obtained in the same way, and converted to 6-dimensional vectors
in the same manner, as the training
examples.  This results in an accuracy score of correctly classified
test examples.
We ran 100 experiments. The actual data are available at
\cite{Ci04}.
\begin{figure}
\centering {\tiny
\includegraphics[angle=-90,width=3in]{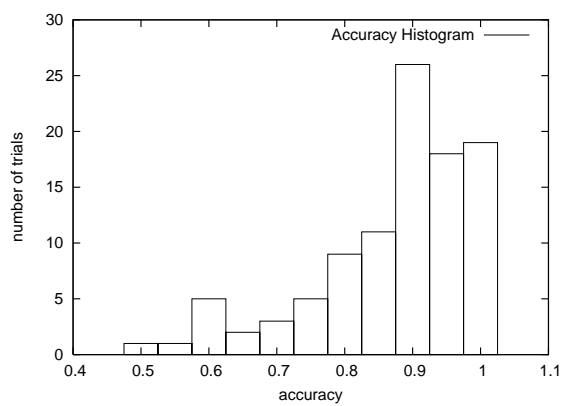}
}
\caption{Histogram of accuracies over 100 trials of WordNet experiment.}
\label{fig.wordnethisto}
\end{figure}
A histogram of agreement accuracies
is shown in Figure~\ref{fig.wordnethisto}.
On average, our method turns out to agree well with the WordNet semantic
concordance made by human experts.  The mean of the accuracies
of agreements is 0.8725.
The variance is $\approx 0.01367$, which gives a standard deviation of
$\approx 0.1169$. Thus, it
is rare to find agreement less than
75\%. The total number of Google searches involved in this
randomized automatic trial is upper bounded by $100 \times 70 \times 6 \times 3 =
126,000$. A considerable savings resulted from the fact
that we can re-use certain google counts.
For every new term, in computing its 6-dimensional vector, the \NGD computed
with respect to the six anchors requires the counts for the anchors which needs to be
computed only once for each experiment, the count of the new term which can be
computed once, and the count of the joint occurrence of the new term and
each of the six anchors, which has to be computed in each case.
Altogether, this gives a total of $6+70+70 \times 6 = 496$ for every experiment,
so $49,600$ google searches for the entire trial.

\end{document}